\documentclass{article}

\usepackage[final,nonatbib]{neurips_2025}

\usepackage[utf8]{inputenc} % allow utf-8 input
\usepackage[T1]{fontenc}    % use 8-bit T1 fonts
\usepackage{hyperref}       % hyperlinks
\usepackage{url}            % simple URL typesetting
\usepackage{booktabs}       % professional-quality tables
\usepackage{amsfonts}       % blackboard math symbols
\usepackage{nicefrac}       % compact symbols for 1/2, etc.
\usepackage{microtype}      % microtypography

\usepackage{graphicx}
\usepackage{algorithm}
\usepackage{algorithmic}
\usepackage{amssymb}
\usepackage{wrapfig}
\usepackage{multirow}
\usepackage[table]{xcolor}
\usepackage{arydshln}
\usepackage{float}
\usepackage{amsmath}
\usepackage{fontawesome}

\definecolor{lightgray}{gray}{0.8}
\definecolor{lightblue}{rgb}{0.21,0.49,0.74}
\definecolor{grayline}{RGB}{128,128,128}
\definecolor{blackline}{RGB}{0,0,0}

\hypersetup{
    colorlinks=true,
    breaklinks=true,
    urlcolor=lightblue,
    linkcolor=red,
    bookmarksopen=false,
    filecolor=black,
    citecolor=lightblue,
    linkbordercolor=lightblue
}

\newcommand{\etal}{\textit{et al.}} 
\newcommand{\ie}{\textit{i.e.}}

\title{\vspace{-0.2cm}
\raisebox{-0.15\height}{\includegraphics[width=0.05\linewidth]{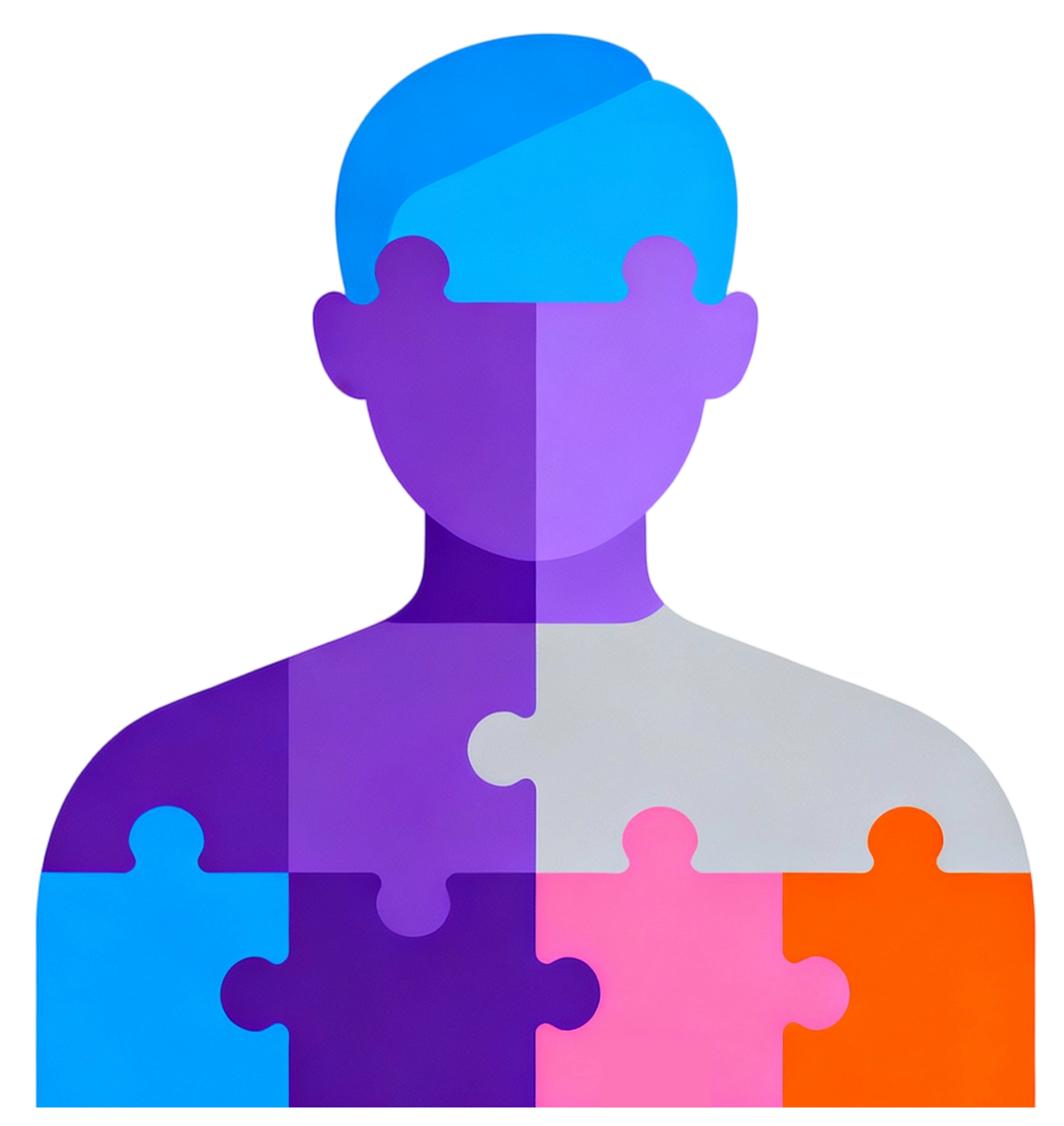}}~ReID5o: Achieving Omni Multi-modal Person \\Re-identification in a Single Model}

\author{%
	\textbf{Jialong Zuo $^{1}$} \quad ~\textbf{Yongtai Deng $^1$} \quad ~\textbf{Mengdan Tan $^1$} \quad ~\textbf{Rui Jin $^1$} \quad ~\textbf{Dongyue Wu $^1$} \\ ~\textbf{Nong Sang $^{1}$} \quad ~\textbf{Liang Pan $^2$} \quad ~\textbf{Changxin Gao $^1$}\thanks{Corresponding Author.}\\
    \\
	$^{1}$ National Key Laboratory of Multispectral Information Intelligent Processing Technology,\\ School of Artificial Intelligence and Automation, Huazhong University of Science and Technology, \\ $^{2}$ Shanghai AI Laboratory.   \\
    {\tt\small\{jlongzuo, cgao\}@hust.edu.cn}
    \\[1ex]
\faGithubAlt~\textbf{Code \& Dataset:} \href{https://github.com/Zplusdragon/ReID5o_ORBench}{\texttt{https://github.com/Zplusdragon/ReID5o\_ORBench}}
}

\begin{document}

\maketitle

\vspace{-8mm}
\begin{abstract}
\vspace{-3mm}
In real-word scenarios, person re-identification (ReID) expects to identify a person-of-interest via the descriptive query, regardless of whether the query is a single modality or a combination of multiple modalities. However, existing methods and datasets remain constrained to limited modalities, failing to meet this requirement. Therefore, we investigate a new challenging problem called Omni Multi-modal Person Re-identification (OM-ReID), which aims to achieve effective retrieval with varying multi-modal queries. To address dataset scarcity, we construct \textbf{ORBench}, the first high-quality multi-modal dataset comprising 1,000 unique identities across five modalities: RGB, infrared, color pencil, sketch, and textual description. This dataset also has significant superiority in terms of diversity, such as the painting perspectives and textual information. It could serve as an ideal platform for follow-up investigations in OM-ReID. Moreover, we propose \textbf{ReID5o}, a novel multi-modal learning framework for person ReID. It enables synergistic fusion and cross-modal alignment of arbitrary modality combinations in a single model, with a unified encoding and multi-expert routing mechanism proposed. Extensive experiments verify the advancement and practicality of our ORBench. A range of models have been compared on it, and our proposed ReID5o gives the best performance.
\end{abstract}

\vspace{-4mm}
\section{Introduction}
\label{sec:intro}
\vspace{-1mm}
Person re-identification (ReID), as a fine-grained retrieval task, aims to search for person images of the target identity based on a given descriptive query~\cite{ye2021deep}. Due to the broad applications of ReID in fields such as urban security, many researchers have conducted in-depth studies on it. It can be classified into single-modal retrieval~\cite{market1501,MSMT17} and cross-modal retrieval~\cite{cuhkpedes,SYSUMM01,pkusketch}. The former involves the mutual search of RGB images, while the latter usually use queries from other modalities to retrieve RGB images. Considering the frequent lack of original RGB queries in real scenarios, the latter has increasingly drawn attention in recent years. 

However, despite the considerable progress in this technology~\cite{PLIP,CION,AIO,InstructReID}, a challenging and practical problem, which aims to achieve effective retrieval with varying multi-modal queries and their combinations in the ReID model, remains largely unexplored in previous researches.

We propose that exploring such a challenging problem, termed Omni Multi-modal Person Re-identification (OM-ReID), is of great significance.
On the one hand, due to the diversity of information acquisition, there is often a need for multi-modal queries in practical applications~\cite{ye2024transformer}. Users expect the model to be able to effectively process multi-modal information and provide comprehensive and accurate retrieval results. If the model can handle these varying modalities, it will significantly boost its practicality.
On the other hand, theoretically, different modalities provide abundant and complementary features~\cite{zhai2022trireid,chen2023modalityagnostic,AIO}. For instance, RGB images are vulnerable to environmental light variations~\cite{SYSUMM01}, and infrared as well as sketch images lack essential color information critical for ReID tasks. Fusing multi-modal features is akin to assembling an information jigsaw from diverse perspectives and dimensions. This enables a more comprehensive portrayal of person characteristics, thereby minimizing misjudgments stemming from single-modal limitations.

\begin{wrapfigure}[22]{r}{6.5cm}
\vspace{-5mm}
    \centering
    \includegraphics[width=\linewidth]{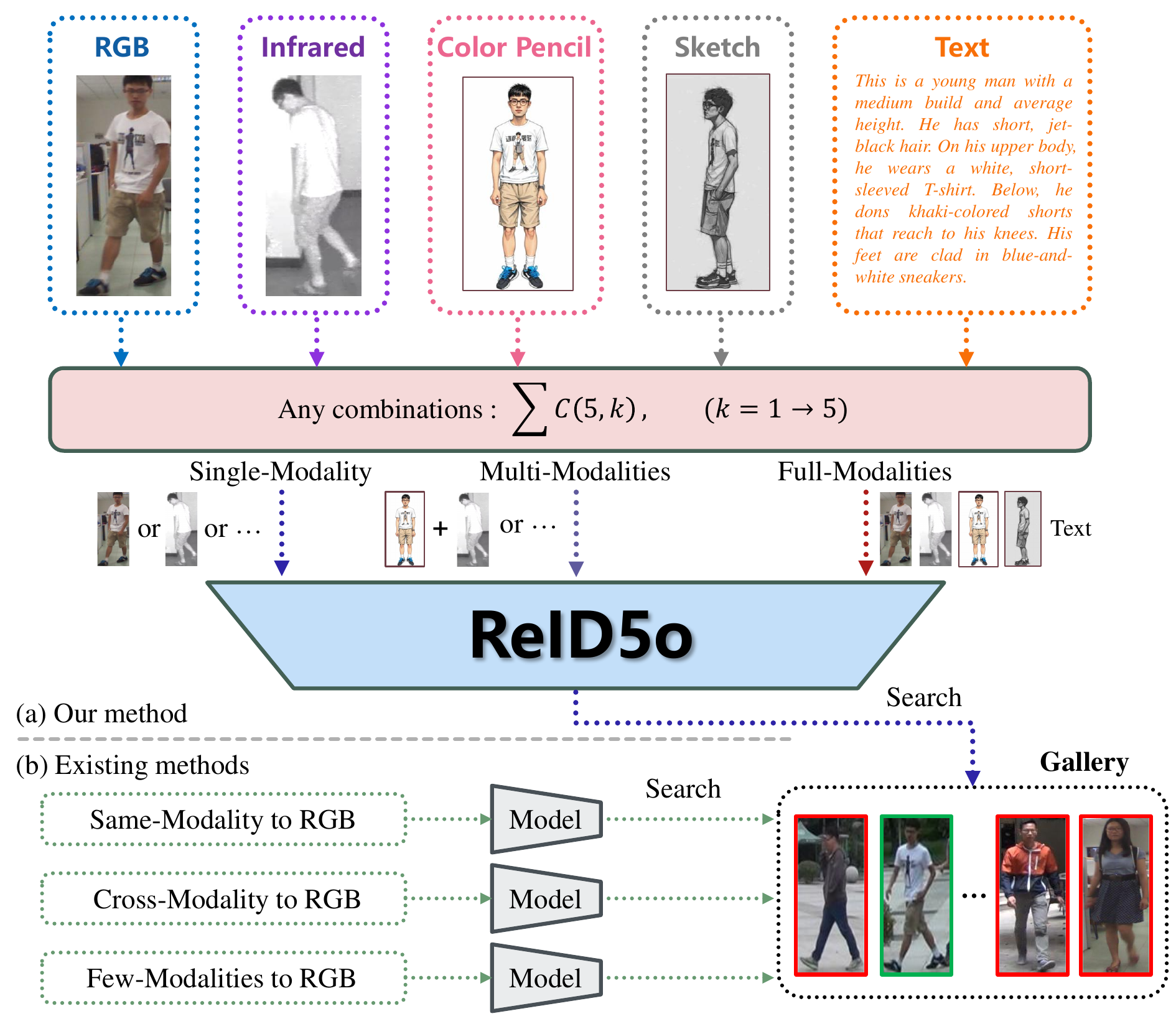}
    \vspace{-7mm}
    \caption{Our \textbf{ReID5o} can effectively conduct retrieval with any combinations of five modalities, adapting to various queries with different uncertain modalities in real scenarios. However, existing methods~\cite{SYSUMM01,cuhkpedes,zhai2022trireid,chen2023modalityagnostic,AIO} are constrained to few modalities and are unable to achieve arbitrary retrieval with five modalities.
    }
    \label{fig:fig1}
    % \vspace{-4mm}
\end{wrapfigure}
Considering the above aspects, and given that the related researches are constrained by datasets involving only few modalities and whose quality is not satisfactory~\cite{market1501,SYSUMM01,cuhkpedes,pkusketch,zhai2022trireid}, this paper makes the following contributions at the levels of both the dataset and the methodology.

\begin{figure}[htb]
    \centering
    \includegraphics[width=\textwidth]{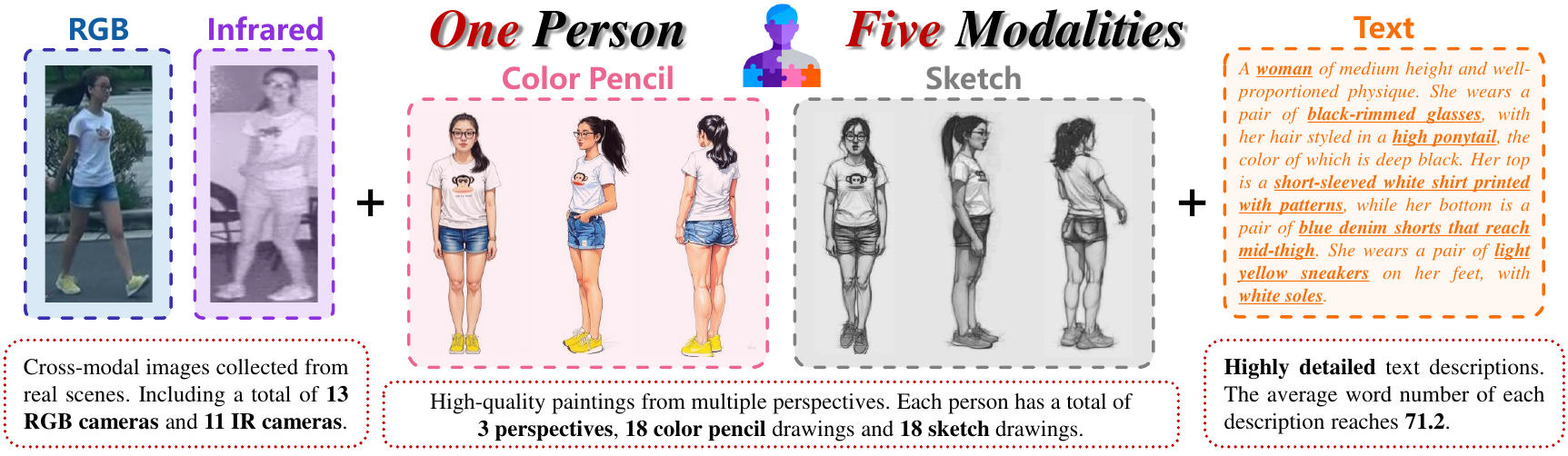}
    \vspace{-6mm}
    \caption{
    \textbf{The overview of our proposed \textbf{ORBench} dataset.} Our dataset is remarkable for containing rich, high-quality and diverse five-modal descriptive data for the same person, offering a comprehensive and in-depth resource for person ReID research.
    }
    \label{fig:fig2}
    \vspace{-4mm}
\end{figure}

We collect the first high-quality person ReID dataset with 5 modalities, named \textbf{ORBench}. As shown in Fig.~\ref{fig:fig2}, for the same person, our dataset simultaneously contains RGB images, infrared images, colored pencil drawings, sketch drawings, and textual descriptions, providing a comprehensive portrayal of the person. Moreover, our dataset exhibits outstanding characteristics in terms of diversity. In addition to the cross-camera perspectives inherent in the real-captured data~\cite{SYSUMM01,LLCM}, the contained paintings simultaneously reflects the frontal, dorsal, and lateral appearance features of the same person. Furthermore, the text component adopts an highly detailed and unrestricted descriptive style. By performing elaborate manual annotation and recurrent corrections, ORBench has significant superiority in terms of quality and diversity, compared to existing datasets. It contains 45,113 RGB images, 26,071 infrared images, 18,000 color pencil drawings, 18,000 sketches and 45,113 textual descriptions of 1,000 identities in total.

In addition, as shown in Fig.~\ref{fig:fig1}, to tackle the new challenge of retrieving with multiple modalities, we further propose a novel unified model named \textbf{ReID5o}, by which the portrayals of the same person across different modalities and their combinations are explicitly correlated to deeply mine the identity-invariance. ReID5o first encodes the inputs from diverse modalities into a shared embedding space via a multi-modal tokenizing assembler. Subsequently, a multi-expert router is devised to facilitate effective modality-specific representation learning within a unified feature extractor. Then, a feature mixture and a simple alignment strategy~\cite{IRRA} are utilized to implement efficient fusion of multi-modal representations and their correlation learning, respectively. Compared with existing methods, ReID5o achieves the best performance and can serve as a simple but effective baseline method for our ORBench.

We conduct comprehensive experiments to verify the advancement and effectiveness of our dataset and method. It can be concluded that multi-modal combination querying brings significant performance improvements, and it serves as a more practical approach in real scenarios. As a research focusing on person ReID with arbitrary modalities and their combinations, we believe that our first high-quality dataset, ORBench, and our method, ReID5o, can inspire the subsequent development in this field.

\vspace{-3mm}
\section{ORBench Dataset}
\vspace{-2mm}
\subsection{Construction Motivation}
\vspace{-1mm}
Multi-modal learning~\cite{wang2023large,zhu2024vision+,li2024multimodal} has become a hot topic in the computer vision field. However, when directing to person ReID, researchers often confine themselves to datasets with few modalities and unsatisfactory quality~\cite{cuhkpedes,pkusketch,zhai2022trireid,chen2023modalityagnostic,AIO}. Therefore, we construct ORBench, the first high-quality dataset with five modalities, driven by the following two major motivations. 

Filling the modal gap and improving data quality. In the field of person ReID, previous datasets only covered few modalities, making it difficult to comprehensively depict the characteristics of persons. In addition, the quality of some multi-modal datasets~\cite{zhai2022trireid,chen2023modalityagnostic} is obviously poor, and only rough generation strategies are utilized to obtain unrealistic multi-modal data. Therefore, to fill such a gap, we are determined to construct a new high-quality dataset that encompasses five modalities in a pioneering way, faithfully covering the characteristics of persons. 

Promoting the development of multi-modal technologies and driving innovation. Multi-modal fusion and retrieval technologies~\cite{zhu2023multi,wang2016comprehensive} have great potential in person ReID and there is an urgent need for their practical applications. However, due to the lack of suitable multi-modal datasets, their development is severely restricted. Therefore, to build an ideal platform for the development of this technology and drive innovation, we construct our ORBench, serving as the first high-quality multi-modal dataset in this field.

\vspace{-2mm}
\subsection{Dataset Collection}
\vspace{-2mm}
Based on the existing RGB-infrared datasets~\cite{SYSUMM01,LLCM}, our ORBench dataset, with the cooperation of manual annotation and image generation experts, achieves the supplementation of additional modalities including color pencil drawings, sketch drawings, and textual descriptions. In addition, a manual correction and review mechanism has been incorporated to ensure the high quality.

For the RGB and infrared data, we manually selected persons with representative appearances from the existing datasets SYSU-MM01~\cite{SYSUMM01} and LLCM~\cite{LLCM}. That is to say, we filtered out those person images with poor imaging conditions and weak clothing representativeness. For example, if the appearance features of a person can hardly be seen in both the dual-modal images, the data of this person will be removed. Eventually, we obtained 45,113 RGBs and 26,071 IRs of 1,000 persons.

For the color pencil data, we utilized an online image generation model~\cite{doubao} as our painting agent, and incorporated extensive manual supervision to ensure the drawing quality. Specifically, for each person, we first selected some multi-perspective RGB images with high quality, and manually annotated them with detailed textual descriptions. Then, we required the agent to refer to these descriptions and images to paint color pencil drawings of the person's front, back, and side views. In addition, with many color pencil drawings painted, we manually selected six most satisfactory drawings for each view, to further ensure the quality. Eventually, we obtained 18,000 color pencil drawings.

For the sketch data, considering the maturity of the sketch synthesis technology, after conducting extensive investigations, we decided to utilize the Meitu application software~\cite{Meitu} to directly convert the obtained colored pencil drawings into realistic sketch drawings. For the text data, we hired a few unique workers to be involved in the text annotation task and instructed them to describe the important characteristics for each RGB person image in detail. Through the above two procedures, we obtained 18,000 sketch drawings and 45,113 textual descriptions in total.

\begin{table}[htb]
\tiny
\centering
\caption{\textbf{Comparisons with other ReID datasets.} ORBench is the first to have rich five-modal data and is the only dataset covering color pencil drawings. Only relatively high-quality datasets with manual annotations are listed, and low-quality synthetic datasets are excluded.}
\vspace{1mm}
\label{table1}
\resizebox{\linewidth}{!}{
\begin{tabular}{lcccccccc}
\hline
\textbf{Datasets}                                                     & \textbf{Year} & \textbf{Modality} & \textbf{\#Identities} & \textbf{\#RGB Imgs} & \textbf{\#IR Imgs} & \textbf{\#CP Imgs} & \textbf{\#SK Imgs} & \textbf{\#Texts} \\ \hline
Market1501~\cite{market1501}                                                            & 2015          & R                 & 1,501                 & 32,668              & -                  & -                  & -                  & -                \\
CUHK-PEDES~\cite{cuhkpedes}                                                            & 2017          & R,T               & 13,003                & 40,206              & -                  & -                  & -                  & 80,412           \\
MSMT17~\cite{MSMT17}                                                                & 2018          & R                 & 4,101                 & 126,441             & -                  & -                  & -                  & -                \\
PKU-Sketch~\cite{pkusketch}                                                            & 2018          & R,S               & 200                   & 400                 & -                  & -                  & 200                & -                \\
SYSU-MM01~\cite{SYSUMM01}                                                             & 2020          & R,I               & 491                   & 30,071             & 15,792             & -                  & -                  & -                \\
ICFG-PEDES~\cite{icfgpedes}                                                            & 2021          & R,T               & 4,102                 & 54,522              & -                  & -                  & -                  & 54,522           \\
TriReID~\cite{zhai2022trireid}                                                            & 2022          & R,S,T               & 200                 & 5,600              & -                  & -                  & 200                  & 5,600           \\
LLCM~\cite{LLCM}                                                                  & 2023          & R,I               & 1,064                 & 25,626              & 21,141             & -                  & -                  & -                \\
MaSk1K~\cite{linDomainGapExploiting2023}                                                                & 2023          & R,S               & 1,501                 & 32,668              & -                  & -                  & 4,763              & -                \\

UFine6926~\cite{zuo2024ufinebench}                                                             & 2024          & R,T               & 6,926                 & 26,206              & -                  & -                  & -                  & 52,412           \\ \hline
\rowcolor{gray!20}\textbf{ORBench}                                    & \textbf{2025}          & \textbf{R,I,C,S,T}         & \textbf{1,000}                 & \textbf{45,113}             & \textbf{26,071}             & \textbf{18,000}             & \textbf{18,000}             & \textbf{45,113}           \\ \hline
\end{tabular}}
\vspace{-3mm}
\end{table}

\vspace{-3mm}
\subsection{Dataset Statistics}
\vspace{-2mm}
Benefiting from our meticulous and labor-intensive manual annotation, ORBench is full of rich and high-quality multi-modal data. Compared with existing ReID datasets, our ORBench enjoys the following advantages: 

\noindent
\textbf{Most Modalities to Date.} Most previous multi-modal datasets~\cite{SYSUMM01,cuhkpedes,pkusketch} have merely two modalities, while for few tri-modal datasets~\cite{zhai2022trireid,chen2023modalityagnostic}, the extra modalities are acquired via rough data synthesis strategies. As shown in Tab.~\ref{table1}, ORBench is the person ReID dataset boasting the greatest number of modalities so far, offering a more comprehensive and detailed portrayal of persons.

\noindent
\textbf{High Quality.} Existing tri-modal datasets often have poor quality due to rough data synthesis. For instance, Tri-CUHK-PEDES~\cite{chen2022sketch} just converts low-res RGB images to sketches, resulting in blurriness, missing parts, and distortion. TriReID~\cite{zhai2022trireid} utilizes limited-capacity caption models for rough texts. However, our ORBench ensures high-quality data through careful, labor-intensive manual annotation. More examples are shown in the supplement.

\noindent
\textbf{Competitive Dataset Scale}. As shown in Tab.~\ref{table1}, compared with existing datasets, the scale of our dataset is highly competitive, containing 45,113 RGB images, 26,071 infrared (IR) images, 18,000 color pencil (CP) drawings, 18,000 sketch (SK) drawing and 45,113 texts of 1,000 persons. Notably, our scale especially outshines those datasets with paintings. For example, compared with the previous largest sketch dataset, MaSk1K~\cite{linDomainGapExploiting2023}, we have approximately 4 times the number of sketches.

\subsection{Evaluation Protocol}
There are 1,000 valid identities in ORBench dataset. We have a fixed split using 600 identities for training and 400 identities for testing. During training, all multi-modal data for the 600 persons in the training set can be applied.

During the testing phase, the samples from the RGB modality are regarded as the gallery set, while the samples from the remaining four modalities and their combinations are treated as the corresponding query sets. Specifically, we design four search modes: single-modal search (MM-1), dual-modal search (MM-2), tri-modal search (MM-3), and quad-modal search (MM-4). For single-modal search, the samples of each modality form their respective query sets. For the remaining three multi-modal searches, we first select a primary modality. Then, for each sample of this modality, we randomly select samples of the same identity from the remaining modalities as supplementary ones to form a query tuple. Totally, there are 4 query sets for MM-1, 12 for MM-2, 12 for MM-3, and 4 for MM-4. The performance of each mode is measured as the average of retrieval results across its query sets.

As a common practice, we adopt Cumulative Matching Characteristic (CMC) and mean average precision (mAP) to evaluate the performance. Given a query, all gallery images are ranked according to their affinities with the query. A successful search is achieved if any image of the corresponding person is among the rank-k images.

\subsection{Discussion on Realism and Idealized Assumptions.}
It is important to acknowledge the idealized assumptions underpinning our dataset construction. We operated under the premise that comprehensive eyewitness information enables the creation of a complete textual description and, subsequently, a high-fidelity visual rendering of the target. Consequently, the identity consistency across modalities in ORBench is intentionally high.

In the taxonomy of~\cite{klare2010matching}, our sketches and color paintings are best characterized as "viewed sketches"—precise depictions derived from ample information—rather than incomplete or inaccurate "forensic sketches" that result from the fallibility of human memory. This design choice was necessitated by the feasibility of large-scale dataset construction. While ORBench provides a foundational, high-quality benchmark for studying multi-modal alignment in a controlled setting, it does not fully capture the noise and variance inherent in real-world forensic scenarios. We posit that this clean dataset serves as a crucial starting point and a powerful pre-training resource.
\vspace{-1mm}
\section{ReID5o: Method}
\vspace{-1mm}

\begin{wrapfigure}[20]{r}{7cm}
\vspace{-13mm}
    \centering
    \includegraphics[width=\linewidth]{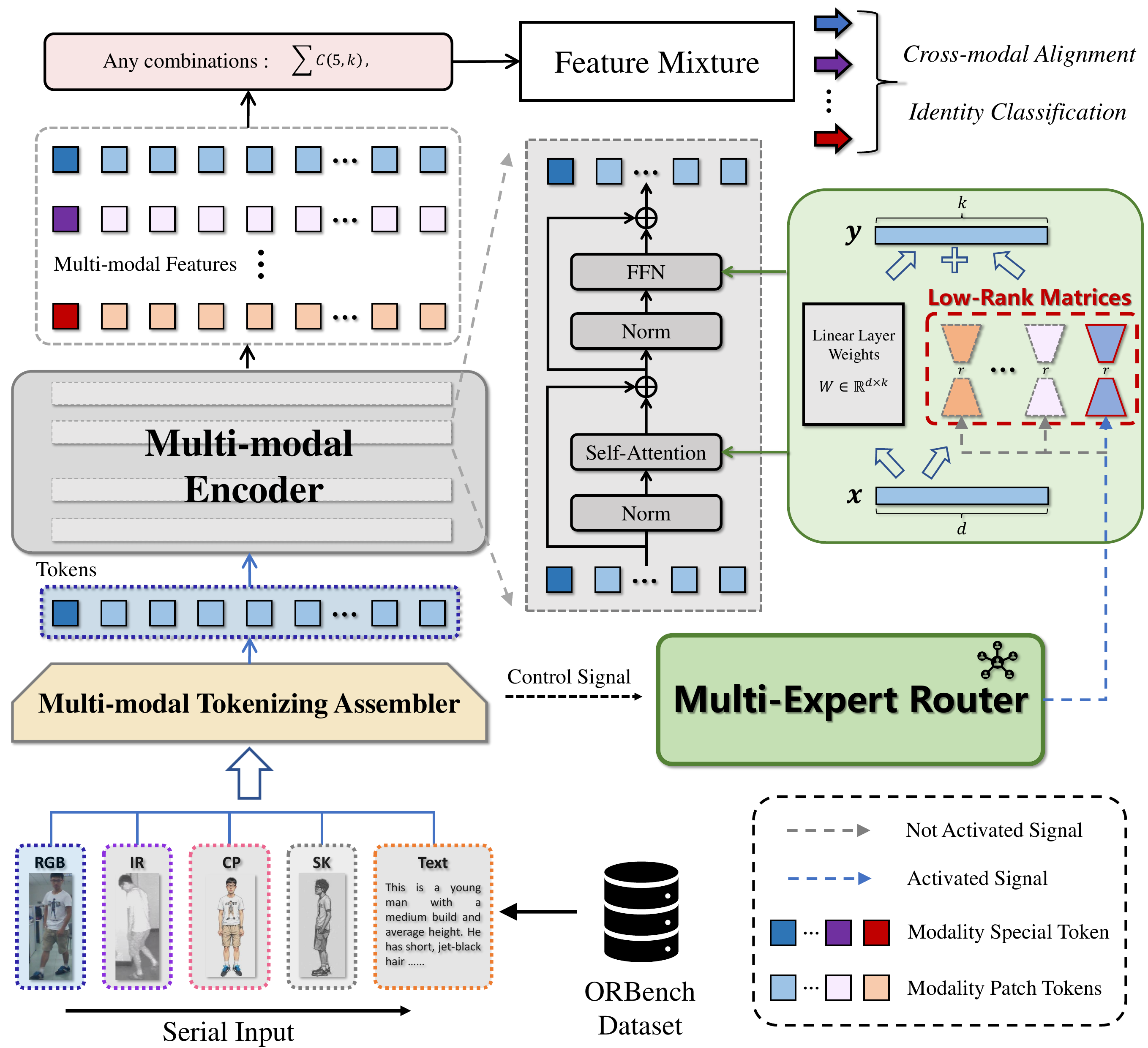}
    \vspace{-6mm}
    \caption{\textbf{The schematic of our proposed ReID5o framework.} As the unified multi-modal encoder extracts the modality-shared features, our specially designed multi-expert router can effectively promote the modality-specific representation learning.}
    \label{fig:fig4}
    \vspace{-3mm}
\end{wrapfigure}

In this section, we propose a unified multi-modal learning framework for person ReID, named ReID5o. We utilize a multi-modal tokenizing assembler to encode the multi-modal inputs into a shared embedding space. Then, a unified encoder is used to extract modality-shared features, and a multi-expert router is designed to facilitate modality-specific representation learning. Also, a feature mixture fuses the multi-modal features. Finally, we use the SDM~\cite{IRRA} and identity classification losses to implement cross-modal alignment. The whole schematic is shown in Fig.~\ref{fig:fig4}.

\subsection{Multi-modal Tokenizing Assembler}
To encode varying-modal data inputs into a shared embedding space, we design a straightforward multi-modal tokenizing assembler. This assembler is composed of five sub-tokenizers, which are respectively used to tokenize the RGB, IR, CP, SK, and text data. In addition, it will also provide discrete control signals for the subsequent multi-expert router according to the modality difference. 

Specifically, for these four types of visual inputs, we use non-shared visual tokenizers~\cite{CLIP} with the same structure to encode them respectively. This visual tokenizer consists of a simple convolutional layer and an LN layer. In addition, we add an additional class token and positional embeddings as a common practice. For text input, following standard convention~\cite{CLIP}, we first add special flags at the beginning and end of the text respectively. Then, we employ a vocabulary mapping layer and positional embeddings to project the input text into a high-dimensional embedding space. Formally, for the input data $\mathbf{X}^{mod}$, the process of obtaining its corresponding embeddings can be formulated as follows. 
\begin{equation}
\mathbf{E}^{mod},c^{mod} = \gamma^{mod}(\mathbf{X}^{mod}),
\end{equation}
where $\mathbf{E}^{mod} \in \mathbb{R}^{n\times D^e}$, representing that the input $\mathbf{X}^{mod}$ is encoded into $n$ tokens with dimension $D^e$. $c^{mod}$ represents a binary signal for modality activation, which is used to control the subsequent multi-expert router. Here, $\gamma^{mod}$ represents the tokenizer of each modality and $mod \in \{R, I, C, S, T\}$, which respectively represent the RGB, Infrared, Color pencil, Sketch, and Text modalities. 

\vspace{-2mm}
\subsection{Multi-Expert Router}
We utilize a unified multi-modal encoder to extract modality-shared features, which inherits the rich multi-modal pre-trained knowledge from CLIP~\cite{CLIP} to ensure a good initial starting point for subsequent training. As a Transformer, it contains multiple linear layers. For a certain linear layer $W \in \mathbb{R}^{d\times k}$, its function can be reduced as $\textbf{y} = W(\textbf{x})$, where $\textbf{x} \in \mathbb{R}^{d}$ and $\textbf{y} \in \mathbb{R}^{k}$ represent the input and output feature for this layer, respectively. It can be seen that if the same linear layer is used directly for different modalities, it is hard to extract modality-specific representations.

Therefore, we propose a simple yet effective multi-expert router to explicitly promote the modality-specific representation learning based on the modality categories. Specifically, we incorporate multiple modality-specific low-rank matrices~\cite{lora} into each linear layer as efficient feature extraction experts. In the absence of data input, these experts stay inactive. Once the assembler encodes data from a particular modality, the router will receive the associated control signal and activates the corresponding expert, enabling modality-specific parameter updates. Formally, for the input feature $\mathbf{x}^{mod}$ of a certain linear layer $W$, its output feature can be formulated as follows.
\begin{equation}
\begin{aligned}
\mathbf{y}^{mod} &= W\mathbf{x}^{mod} + c^{mod}\Delta{W^{mod}}\mathbf{x}^{mod} \\
&= W\mathbf{x}^{mod} + c^{mod}B^{mod}A^{mod}\mathbf{x}^{mod},
\end{aligned}
\end{equation}
where $B \in \mathbb{R}^{d\times r}, A \in \mathbb{R}^{r\times k}$ and $ \ r \ll\min(d,k)$. $c^{mod}$ is a binary control signal from the assembler, indicating whether the expert $\Delta{W^{mod}}$ is activated. We can observe that for any modal input $\mathbf{x}^{mod}$, while $W$ is used to extract the modal shared features, $\Delta{W^{mod}}$ will be employed to extract the modal unique features. In this way, we can effectively uncover the modal invariance and discrimination.

\subsection{Feature Mixture and Learning Strategy}
Now, for the input data $\mathbf{X}^{mod}$, through the aforementioned tokenization and expert routing mechanisms, we can obtain its output features $\mathbf{Z}^{mod} \in \mathbb{R}^{n\times D}$. Then, to achieve full complementary information interaction between different modalities, we employ an efficient feature mixture to fuse the features of varying modality combinations. The feature mixture consists of a multi-head self attention (MSA) layer, 1-layer transformer block and a multi-layer perception (MLP). Then, the combinatorial traversal is performed on the entire set of multimodal query features $\{Z^I,Z^C,Z^S,Z^T\}$. We generate all possible combinations from single-modal to quad-modal. These combinations are then concatenated and processed by the Feature Mixture module.

Taking the combination of $\mathbf{Z}^{I}$ and $\mathbf{Z}^{T}$ as an example, the feature fusing process can be formulated as follows.
\begin{equation}
\mathbf{z}^{I,T} = MLP(TF(MSA(LN(CAT(\mathbf{Z}^I,\mathbf{Z}^T))))),
\end{equation}
where $\mathbf{z}^{I,T}\in\mathbb{R}^{D}$, indicating the IR and Text fused representation. $CAT(\cdot)$, $LN(\cdot)$ and $TF(\cdot)$ denote the concatenation, layer normalization and transformer block, respectively. The last layer of the MLP is an average pooling layer.

In addition, we directly regard the output of special token as the single-modal representation $\mathbf{z}^{mod}$. As a simple learning strategy, we adopt the commonly used SDM loss~\cite{IRRA} and identity classification (IC) loss to supervise the learning process. Considering that RGB modality is usually utilized as the gallery set and contains more original and complete information, we take it as the core target for feature alignment. We explicitly conduct the alignment of diverse modality combinations. To realize this, the overall learning objective can be written as follows.
\begin{equation}
\mathcal{L} = \sum_{R \notin c_i} SDM(\mathbf{z}^{R},\mathbf{z}^{c_i}) + \alpha\sum_{c_i} IC(\mathbf{z}^{c_i}), 
\end{equation}
where $c_i \in \{A|A\subseteq S,A\neq\varnothing\}$ and $S = \{R, I, C, S, T\}$. $\alpha$ is a manually set fixed hyper-parameter, which is used to control the importance of the identity classification loss.

\vspace{-2mm}
\section{Experiments}
\vspace{-2mm}
There is no existing person ReID methods specifically designed to achieve the retrieval of any combinations of these five modalities. We investigate a wide range of possible solutions based on some leading general multi-modal models~\cite{CLIP,MetaTF,ImageBind} and cross-modal person ReID models~\cite{chen2023modalityagnostic,PLIP,AIO,IRRA,RDE}, and compare these solutions with our proposed method. We also conduct experiments to verify the data quality of our proposed ORBench dataset. Extensive experiments and comparisons demonstrate the superiority of our ORBench dataset and ReID5o method.

\vspace{-2mm}
\subsection{Experimental Settings}
\vspace{-1mm}
\noindent
\textbf{Datasets.} The main experiments are conducted on our proposed multi-modal ORBench dataset. This dataset is divided into a training set with 600 persons and a testing set with 400 persons. If there is no special statement, the performance is evaluated with the four proposed search modes. In addition, CUHK-PEDES~\cite{cuhkpedes}, ICFG-PEDES~\cite{icfgpedes} and UFine6926~\cite{zuo2024ufinebench} are used for comparisons.

\noindent
\textbf{Implementation Details.} We employ a pre-trained multi-modal encoder~\cite{CLIP}, \ie, CLIP-B/16, as the unified encoder, which is pre-trained on billions of image-text pairs with contrastive learning. In the actual implementation, to fully leverage pre-trained knowledge, we employ an independent text encoder for the textual modality, while adopting the aforementioned strategy for other visual modalities. For the multi-modal tokenizing assembler (MTA), the tokenizers for all modalities are initialized simultaneously with the pre-trained parameters in CLIP~\cite{CLIP}, and will not be shared during the subsequent training. For the multi-expert router (MER), each modality will be assigned an independent expert, and the low-rank matrices~\cite{lora} of a certain expert will be installed on each linear layer of the encoder, where $r$ is set to 4.
For each layer of the feature mixture (FM), the hidden size and number of heads are set to 512 and 8. The hyper-parameter $\alpha$ for the ID loss is set to 1.0. During training, all images are uniformly resized to $384\times 128$ and the maximum length of the textual tokens is set to 77. Also, random erasing, horizontally flipping and crop with padding are employed for image augmentation. Random masking and replacement is employed for text augmentation. Our ReID5o is trained with Adam~\cite{kingma2014adam} for 60 epochs with an initial learning rate $1e^{-5}$. We spend 5 warm-up epochs linearly increasing the learning rate from $1e^{-6}$ to $1e^{-5}$. For the random-initialized experts and feature mixture, the initial learning rate is set to $5e^{-5}$. We use a single A100 80GB GPU.

\vspace{-3mm}
\subsection{Ablation Study and Analysis}
\vspace{-1mm}

\begin{wraptable}[9]{r}{0.5\textwidth}
\vspace{-7mm}
\centering
\caption{\textbf{Ablation study on each component of ReID5o.} We report the mAP results for each search mode, with the best in bold.}
\vspace{1mm}
\resizebox{\linewidth}{!}{
\begin{tabular}{c|ccc|cccc}
\hline 
\multicolumn{1}{c|}{\multirow{2}{*}{No.}}  & \multicolumn{3}{c|}{Components}&\multicolumn{4}{c}{ORBench}\\
\cline{2-8}
& MTA & MER &FM&MM-1&MM-2&MM-3&MM-4\\
\hline 
0 &&& &49.24	&66.36&74.21&78.42	\\
1&\checkmark&& &51.85	&67.43&75.43&79.81	\\
2 &\checkmark&\checkmark& &56.99	&73.58&81.42&85.14	\\
3&\checkmark&& \checkmark&52.80	&69.23&77.70&81.87\\
\rowcolor{gray!20}
4&\checkmark&\checkmark&\checkmark&\textbf{58.09}	&\textbf{75.26}&\textbf{82.83}&\textbf{86.35}	\\
\hline
\end{tabular}}
\label{ablation}
% \vspace{-1mm}
\end{wraptable}
\textbf{Effectiveness of Each Component.} To verify the contribution of each component in ReID5o, we conduct an ablation experiment on our ORBench. The results are reported in Tab.~\ref{ablation}. In the baseline No.0, all visual modalities use the same shared visual tokenizer, and there is no expert router to extract modality-specific features. In addition, simple average pooling is employed for multi-modal feature fusion. As evident from Tab.~\ref{ablation}, compared with the baseine method, each introduced component proves crucial for the overall performance of our ReID5o model, and combining all of them leads to the best performance. 

\begin{wraptable}[10]{r}{0.5\textwidth}
\vspace{-6mm}
\centering
\caption{\textbf{Different settings of the  multi-expert router.} We report the mAP results for each search mode, with the best in bold.}
\vspace{1mm}
\resizebox{\linewidth}{!}{
\begin{tabular}{lcc|cccc}
\hline 
\multicolumn{1}{c}{\multirow{2}{*}{Method}} & \multicolumn{1}{c}{Params.}& \multicolumn{1}{c|}{FLOPs.}&\multicolumn{4}{c}{ORBench}\\
\cline{4-7}
& (M) & (G) &MM-1&MM-2&MM-3&MM-4\\
\hline 
w.t MER &-&-&52.80	&69.23&77.70&81.87\\
\rowcolor{gray!20}
$r=4$&2.36&3.64&\textbf{58.09}	&\textbf{75.26}&\textbf{82.83}&\textbf{86.35}\\
$r=8$ &4.72&7.29&57.59	&74.59&82.33&85.71\\
$r=16$&9.44& 14.57&57.81	&74.90&82.59&86.21\\
$r=32$&18.87&29.14& 57.56	&74.89&82.39&85.72	\\
w. UE&256.4& -&52.42	&69.13&76.36&80.15\\
\hline
\end{tabular}}
\label{ablation_MER}
% \vspace{-1mm}
\end{wraptable}
\textbf{Analysis of the Multi-Expert Router.} To analyze the effectiveness and efficiency of the multi-expert router, we conduct ablation experiments on the setting of $r$ in the low-rank matrices. In Tab.~\ref{ablation_MER}, we present the performance of each setting, as well as the introduced additional parameters and FLOPs compared to the baseline method (w.t. MER). We also analyze a simple and straightforward approach, that is, utilizing unique encoders to encode each modality separately (w. UE). We can observe that UE fails to bring significant improvement while introducing extremely large additional parameters. However, when $r = 4$, our multi-expert router achieves the best performance, and maintains a relatively low additional computational cost.

\begin{wraptable}[12]{r}{0.5\textwidth}
\vspace{-7mm}
\tiny
\centering
\caption{\textbf{Queries with different modality combinations.} The text modality is the primary modality. The best results are in bold.}
\vspace{1mm}
\resizebox{\linewidth}{!}{
\begin{tabular}{l|cccc}
\hline 
\multicolumn{1}{c|}{\multirow{2}{*}{Modality}} &\multicolumn{4}{c}{ORBench}\\
\cline{2-5}
 &Rank-1&Rank-5&Rank-10&mAP\\
\hline 
T &63.15	&77.94&82.92&54.88\\
T+I&81.19& 92.19 & 95.19 & 70.77\\
T+C &91.01 & 96.83 & 98.04 & 81.14\\
T+S&84.71 &93.39 & 95.55 & 74.01\\
T+I+C&95.85 & 99.25 & 99.68 & 86.75\\
T+I+S&92.85 & 98.01 & 98.94 & 81.98\\
T+C+S&93.19 & 97.80 & 98.83 & 83.05\\
\rowcolor{gray!20}
T+I+C+S&\textbf{96.79} & \textbf{99.37}  &\textbf{99.78} & \textbf{87.46}\\
\hline
\end{tabular}}
% \vspace{-3mm}
\label{ablation_MC}
\end{wraptable}
\textbf{Analysis of Multi-modal Complementarity.} Taking the text modality as the primary modality, we explore the impact of the introduction of different modalities on its retrieval results, so as to investigate the complementarity among multi-modal data. As shown in Tab.~\ref{ablation_MC}, the introduction of any additional modality into the query can directly enhance the retrieval performance. Among them, the promoting effect of the C modality is the most obvious. Compared with the T modality, when using T and C modalities, the mAP has increased by 26.26\%. In addition, regarding the C modality, the complementarity effect of the I modality is significantly better than that of the S modality. Compared with T + C, T + I + C can bring about a 5.61\% increase in mAP, while T +  C +S only brings a 1.91\% increase. 

\begin{wraptable}[10]{r}{0.5\textwidth}
\vspace{-6mm}
\tiny
\centering
\caption{\textbf{Ablation study on different learning strategies.} We report the mAP results for each search mode, with the best in bold.}
\vspace{1mm}
\resizebox{\linewidth}{!}{
\begin{tabular}{l|cccc}
\hline 
\multicolumn{1}{c|}{\multirow{2}{*}{Strategy}} &\multicolumn{4}{c}{ORBench}\\
\cline{2-5}
 &MM-1&MM-2&MM-3&MM-4\\
\hline 
ITC~\cite{CLIP} & 53.44	&67.35&74.62&78.25	\\
SupITC~\cite{SUPITC}& 48.16	&64.84&73.03&77.34	\\
CMPM~\cite{CMPM} & 48.02	&67.27&75.41&79.51	\\
SDM~\cite{IRRA}& 57.65	&74.51&82.14&85.60	\\
\rowcolor{gray!20}
SDM+IC~\cite{IRRA}&\textbf{58.09}	&\textbf{75.26}&\textbf{82.83}&\textbf{86.35}	\\
\hline
\end{tabular}}
% \vspace{-3mm}
\label{ablation_lr}
\end{wraptable}
\textbf{Different Learning Strategies.} 
The different implementations of the loss functions for the learning strategy are shown in Tab.~\ref{ablation_lr}. We have implemented one instance-level loss function~\cite{CLIP} and three identity-level loss functions~\cite{SUPITC,CMPM,IRRA} for the cross-modal alignment. As we can see, compared with other losses, the SDM loss~\cite{IRRA} leads to the best performance. In addition, introducing the identity classification (IC) loss will further improve the model's performance to 58.09\% mAP for the single-modal search mode.

\subsection{Comparison with Existing Methods}
As shown in Tab.~\ref{tab:results}, we compared the performance of extensive possible methods on our ORBench. These methods can be divided into multi-modal methods in the general domain and cross-modal methods in the field of person ReID. The former usually aligns data from multiple modalities into the same feature space, such as images, audio, videos, and texts. For the former methods, we treat each visual modality in ORBench as a unique image modality for them. That is, we will utilize four different image tokenizers and then perform training on ORBench. For the latter, since most methods only achieve image-text alignment, for this type of methods, we directly regard the visual data of the four modalities as the image modality of these methods. However, for the AIO method~\cite{AIO} that achieves alignment among RGB, IR, SK, and Text, we additionally add a CP tokenizer to its framework and conduct fine-tuning on our ORBench using its default settings. In addition, since none of these methods achieve explicit alignment between any modality combinations and RGB, we implement an equivalent multi-modal search mode by superimposing the similarity lists of queries and gallery for each modality. We can observe that, compared with the existing methods, our ReID5o achieves the best performance in all four search modes. These results validate the effectiveness of our method in leveraging arbitrary modality combinations for person ReID, and it can serve as the first strong baseline method for our ORBench.

\begin{table*}[t!]
\small
    \centering
    \caption{\textbf{Performance comparison with existing methods on ORBench.} These methods can be divided into cross-modal methods and multi-modal methods. The former performs alignment between two modalities, while the latter performs alignment among multiple modalities. Each method has been fine-tuned on our ORBench. The results* are the outcomes of our reproduction efforts.} 
    \vspace{1mm}
    \resizebox{\textwidth}{!}{
    \begin{tabular}{lcccccccccccc}
    \toprule
        \multirow{2}{*}{\textbf{Method}} & \multicolumn{3}{c}{\textbf{Single-modal Search}}  & \multicolumn{3}{c}{\textbf{Dual-modal Search}} & \multicolumn{3}{c}{\textbf{Tri-modal Search}} & \multicolumn{3}{c}{\textbf{Quad-modal Search}} \\ [0.2em]
        &  {R@1} & {R@10} & {mAP} & {R@1} & {R@10} & {mAP}& {R@1} & {R@10} & {mAP}& {R@1} & {R@10} & {mAP}\\ 
        
        \midrule
        \multicolumn{4}{l}{\textit{Cross-modal Models}} \\
        CLIP~\cite{CLIP}&59.84&77.88&48.73 &80.31&92.80&65.79 &87.67&96.78&73.17 &90.71 & 98.25 &76.69 \\
        PLIP~\cite{PLIP}&60.57&78.24&48.79 &80.78&92.82&65.63 &88.14&96.98&74.60 &90.95&98.12&77.87 \\
        IRRA~\cite{IRRA}&61.17&79.30&51.42 &82.50&94.17&68.83 &89.70&97.39&75.86 &92.65&98.49&79.15 \\
        RDE~\cite{RDE}&60.91&79.20&48.82 &82.36&93.95&67.52 &89.94&97.50&75.27 &92.75&98.63&78.64 \\
        
        \midrule
        \multicolumn{4}{l}{\textit{Multi-modal Models}} \\
        Meta-TF*~\cite{MetaTF}&6.25&19.46&4.24 &10.81&31.26&6.68 &14.82&40.45&8.65 &18.32&47.66&10.45 \\
        ImageBind*~\cite{ImageBind}&60.18&78.32&48.61 &81.78&93.18&67.02 &88.73&97.23&74.20 &91.05&98.36&77.36 \\
        UNIReID~\cite{chen2023modalityagnostic}&59.49&77.59&47.98 &80.91&93.16&66.83 &88.71&97.08&74.86 &91.34&98.02&78.12 \\
        AIO*~\cite{AIO}&50.44&70.88&40.14 &58.82&81.67&49.34 &69.72&89.34&58.58 &75.33&92.52&63.48 \\
        \rowcolor{gray!20}
        \textbf{ReID5o}&\textbf{68.12}&\textbf{86.36}&\textbf{58.09} &\textbf{86.15}&\textbf{96.76}&\textbf{75.26} &\textbf{93.53}&\textbf{99.15}&\textbf{82.83} &\textbf{96.25}&\textbf{99.70}&\textbf{86.35} \\
    \bottomrule
    \end{tabular}}
    \label{tab:results}
    \vspace{-1mm}
\end{table*}

\begin{figure}
\centering
\includegraphics[width=\linewidth]{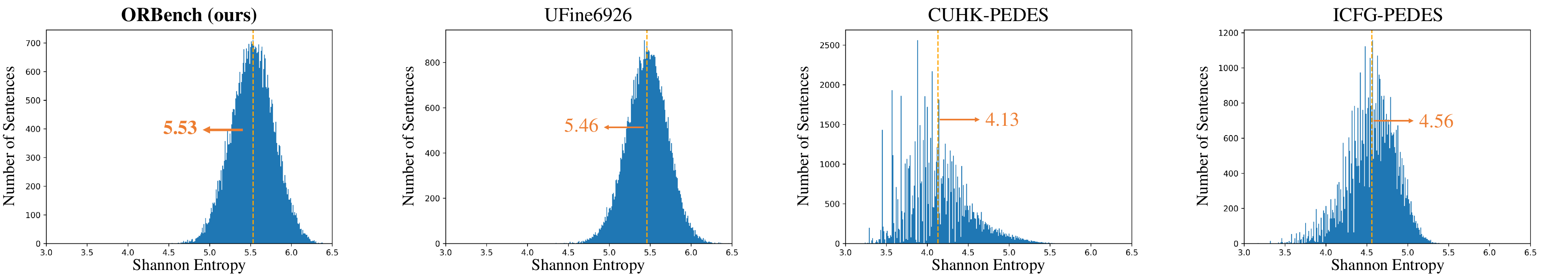}
\vspace{-5mm}
\caption{Comparisons of the Shannon entropy per textual description with existing person ReID datasets containing the text modality. The average entropy of our dataset reaches 5.53, representing the highest level of textual information richness among current datasets.
}
\label{fig:fig5}
\vspace{-3mm}
\end{figure}

\vspace{-2mm}
\subsection{Dataset Quality Assessment}
\vspace{-1mm}
\begin{wrapfigure}[11]{r}{8cm}
\vspace{-8mm}
\centering
\includegraphics[width=\linewidth]{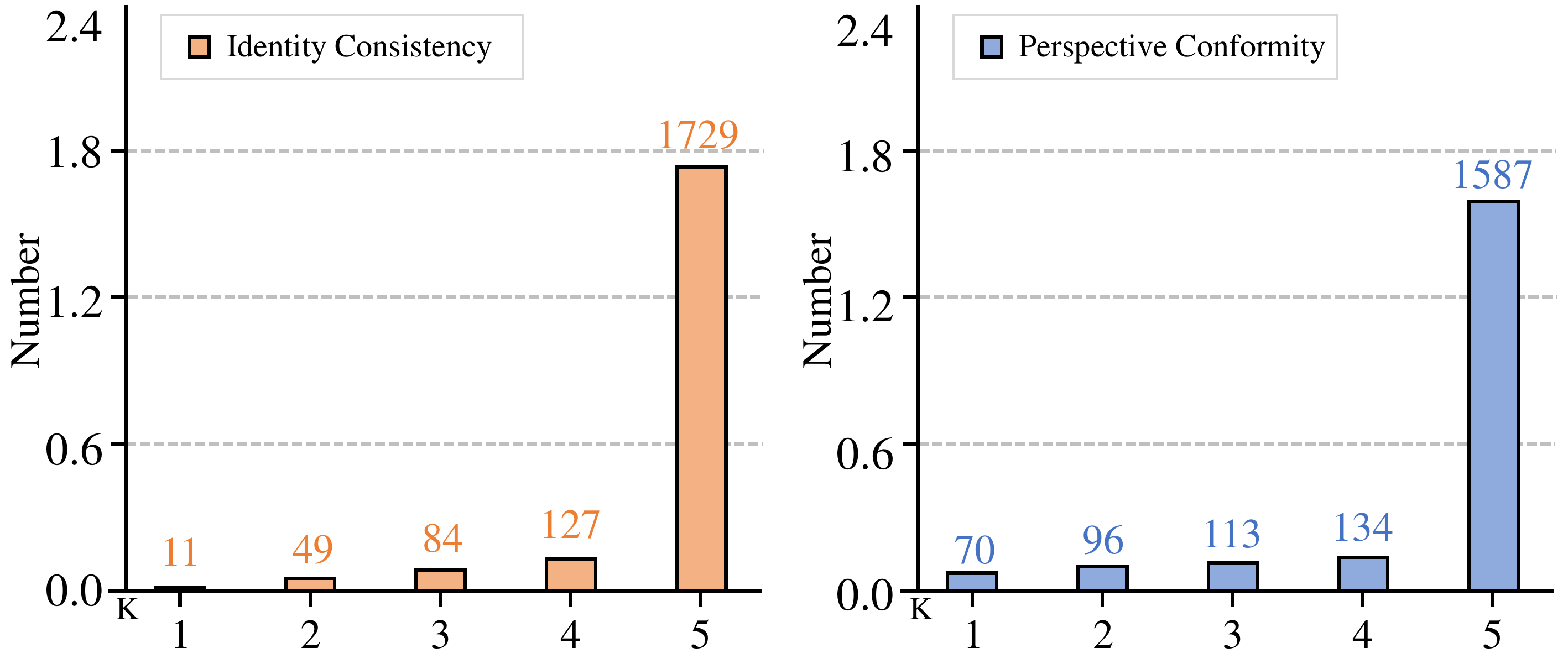}
\vspace{-7mm}
\caption{Public evaluation to assess the identity consistency and perspective comformity of the color pencil drawings in ORBench. 
}
\label{fig:fig6}
\vspace{-1mm}
\end{wrapfigure}
Considering that we additionally annotate data of three modalities to the existing RGB-infrared datasets, and the sketch data is converted from the color pencil data, we conduct experiments to explore the data quality of the text modality and the color pencil modality. 
For the text modality, we use Shannon entropy to measure the amount of information contained in the texts. Specifically, we can treat each word as an event, calculate the probability of each word occurring, and finally calculate the entropy of the entire text. We conduct a statistical comparison of the Shannon entropy per text in our ORBench with those in other datasets~\cite{zuo2024ufinebench,cuhkpedes,icfgpedes}, as illustrated in Fig~\ref{fig:fig5}. We can see that our ORBench has richer textual information than other datasets, verifying the high quality of the text modality. 
For the color pencil modality, we adopt the method of public evaluation to assess the identity consistency and perspective conformity of the color pencil drawings. Specifically, we randomly pick 2,000 samples from the color pencil drawings and split them into 20 groups. Then, 20 evaluators are asked to objectively score these samples on identity consistency with the original RGB images and conformity of drawing perspectives. Higher scores mean better evaluations. Detailed evaluation processes can be found in the supplement. As shown in Fig~\ref{fig:fig6}, the overall quality of the color pencil modality is quite high.

\subsection{Dataset Generalization Evaluation}
\begin{wraptable}[14]{r}{9cm}
\tiny
\vspace{-7mm}
\tabcolsep=3.2pt
\caption{\textbf{Evaluation comparison on dataset generalization.} ``T'', ``C'',``I'',``P",``A'',``O'' denote Tri-CUHK-PEDES~\cite{chen2023modalityagnostic}, CUHK-PEDES~\cite{cuhkpedes}, ICFG-PEDES~\cite{icfgpedes}, PKU-Sketch~\cite{pkusketch}, AIO~\cite{AIO} and our ORBench, respectively. The target datasets are all manually annotated real-world datasets. We report the rank-1 results}
\vspace{1mm}
\label{generalization}
\resizebox{\linewidth}{!}{
\begin{tabular}{ccccccc}
\hline
{Method} & $O{\rightarrow}I$&$T{\rightarrow}I$&$C{\rightarrow}I$&$O{\rightarrow}P$&$T{\rightarrow}P$&$A{\rightarrow}P$\\
\hline
\multicolumn{4}{l}{\textit{Cross-modal Models}} \\
{PLIP~\cite{PLIP}} &\textbf{52.61} &49.23 &49.23&- &- &-\\

{IRRA~\cite{IRRA}} &\textbf{38.52} &32.57 &32.57&- &- &-\\ 

{SketchTrans~\cite{chen2022sketch}} &- &- &-&\textbf{70.30} &66.30 &62.70\\
\hline
\multicolumn{4}{l}{\textit{Multi-modal Models}} \\
{UNIReID~\cite{chen2023modalityagnostic}} &\textbf{36.62} &33.64 &-&\textbf{77.80} &69.80 &65.40\\
{ImageBind~\cite{ImageBind}} &\textbf{38.58} &35.33 &-&\textbf{75.40} &68.60 &61.80\\
{ReID5o} &\textbf{43.27} &38.65 &-&\textbf{80.20} &71.80 &67.00\\
\hline
\end{tabular}}
\end{wraptable}
We conduct experiments to compare the generalization performance of various models trained on different existing multi-modal datasets to other manually annotated datasets. The subtasks are divided into text-based retrieval and sketch-based retrieval. For the former, we use CUHK-PEDES~\cite{cuhkpedes} as the target dataset. For the latter, we used PKU-Sketch~\cite{pkusketch} as the target dataset. To ensure a fair comparison, we employ the same number of training samples for each training dataset, i.e., 40,000 training samples are randomly selected from each dataset. As the results shown in Tab.~\ref{generalization}, for all models, compared with other multi-modal datasets, training on our ORBench dataset can achieve better generalization performance. This strongly demonstrates the high quality of our dataset and its generalizability to real-world application scenarios. Researchers can have full confidence in using our high-quality ORBench dataset.

\vspace{-1mm}
\subsection{Visualization}
\vspace{-1mm}
\begin{wrapfigure}[14]{r}{8cm}
\vspace{-5mm}
\centering
\includegraphics[width=\linewidth]{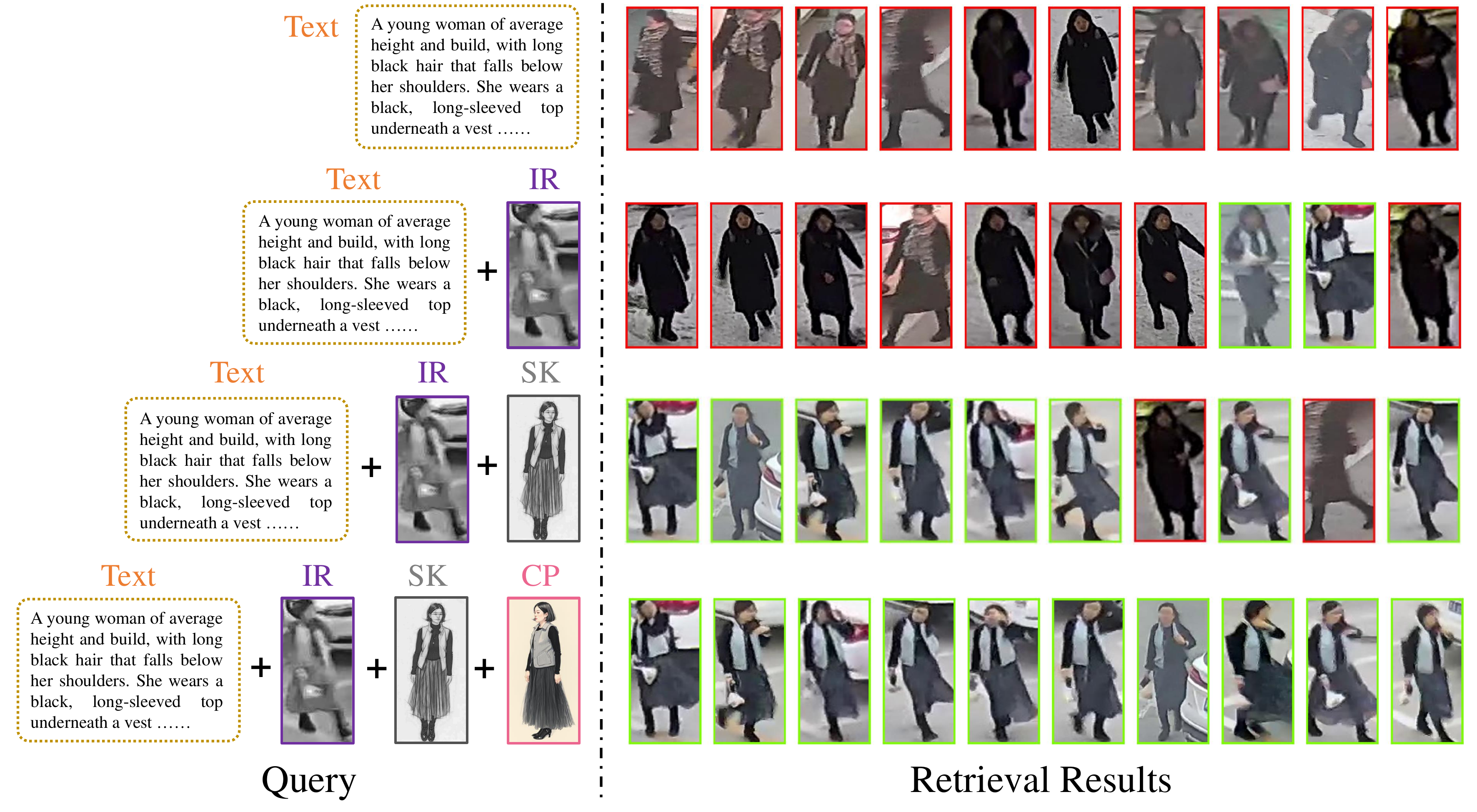}
\vspace{-6mm}
\caption{Visualization of the top-10 ranking lists for four queries with different modalities on ORBench.
}
\label{fig:fig7}
\vspace{-1mm}
\end{wrapfigure}
To demonstrate the role of multi-modal queries more clearly, we visualize retrieval results on ORBench using diverse modality combinations as queries, ranging from single-modal to multi-modal inputs. As shown in Fig.~\ref{fig:fig7}, the addition of supplementary modalities—such as integrating text with infrared or sketch data—significantly improves retrieval accuracy, with top-ranked results becoming more relevant. This highlights that multi-modal queries, by capturing complementary information (e.g., semantic descriptions and visual details under varied conditions), form a more complete depiction of search targets. For practical applications, this implies that leveraging multi-modal descriptive queries—rather than relying on single modalities—can enhance retrieval precision, enabling more accurate identification of targets in complex scenarios.

\section{Related Work}
\label{sec:related_work}
\textbf{Multi-modal Learning.} The goal of multimodal learning is to fully integrate heterogeneous information from multiple modalities to obtain more representative representations of features \cite{liang2023foundations}. With the popularity of Transformers, researchers have proposed unified architectures to process multimodal inputs end-to-end \cite{kim2021vilt,gabeur2020multimodal,li2021align}. Although most models \cite{Botach2022Endtoend, Wang2022BEVT, Poklukar2022Geometric} do not face the problem of missing modalities, this problem frequently occurs in real-world applications. Recent works \cite{Cai2018Deep, Pan2021DiseaseImageSpecific, Wang2023Multimodal, ImageBind, Ma2021SMIL} have introduced various strategies to mitigate this impact. Zhang \etal \cite{ZhangLearningUnseenModality2023} align information from different modalities by mapping them into a shared feature space. Lee \etal \cite{Lee2023CVPR} employ missing-modality-aware prompt learning to  deal with modality absence. Lian \etal \cite{Lian2023GCNet} propose a graph-completion network to jointly optimize classification and reconstruction tasks in an end-to-end framework.

\textbf{Cross-modal Person ReID.} Compared to single-modal person ReID tasks \cite{li2022pyramidal,luo2019bag,zhang2021explainable,chen2023ccsd,hong2025spatial}, cross-modal ReID tasks \cite{chen2023modalityagnostic,SYSUMM01} typically use RGB images as retrieval targets while employing text descriptions \cite{liPersonSearchNatural2017,shao2022learning,gao2021contextual,yang2023unified,li2023clipreid}, infrared images \cite{qiuHighOrderStructureBased2023,ye2022dynamic,yang2023translation}, attributes \cite{wangTransferableJointAttributeIdentity2018}, or sketches \cite{gui2020learning,chen2022sketch} as queries to simulate real-world scenarios with uncertain query modalities. Li \etal \cite{liPersonSearchNatural2017} pioneered text-based person ReID. Wang \etal \cite{wangTransferableJointAttributeIdentity2018} proposed a joint attribute-identity method to learn attribute-semantic and identity-discriminative features. Qiu \etal \cite{qiuHighOrderStructureBased2023} introduced a high-order structure-based intermediate feature learning network to fully exploit structural information in IR features. Lin \etal \cite{linDomainGapExploiting2023} designed mechanisms to mitigate the impact of sketches painter's subjectivity on ReID performance. Subsequent work combines some modalities to address practical constraints in obtaining target modalities and using complementary information. Chen \etal \cite{chen2023modalityagnostic} fused features of RGB, sketch, and text modalities for modality-agnostic retrieval. Li \etal \cite{liAllOneFramework2024} developed a unified framework handling RGB, infrared, sketch, and text modalities. However, existing works are still constraint to scenarios with few modalities and overlook the rich cross-domain information contained in different modality combinations.

\section{Limitation}
\label{limitation}
While the proposed ReID5o framework and the ORBench dataset establish a strong foundation for OM-ReID research, we acknowledge several limitations that point to future directions. First and foremost, ORBench is constructed under an idealized data generation assumption. The high fidelity of its sketch and color painting queries, which resemble "viewed sketches" rather than imperfect "forensic sketches," ensures tight cross-modal alignment but does not fully encapsulate the noise and variance (e.g., inaccuracies, omissions) prevalent in real-world scenarios. Consequently, while our model demonstrates robust performance on this clean benchmark and shows promising generalization in cross-dataset tests, its performance in truly noisy, in-the-wild settings requires further investigation. Secondly, our work is currently limited to a predefined set of five modalities. Other practically relevant modalities, such as thermal infrared or radar, are not included, and our framework is not yet designed for incremental learning of new modalities. Finally, the study focuses on a closed-set identification task, leaving open-world scenarios with unseen identities as an open challenge. We believe addressing these limitations—by introducing realistic noise, extending modality coverage, and moving towards open-set recognition—will be crucial steps for transitioning OM-ReID from a controlled benchmark to robust real-world applications.

\section{Conclusion}

This paper investigated the problem of Omni Multi-modal Person Re-identification (OM-ReID), where the goal is to retrieve a person using a query from any single modality or their arbitrary combinations. To support this research, we constructed ORBench, the first comprehensive dataset for OM-ReID, featuring 1,000 identities across five modalities (RGB, infrared, sketch, color pencil, and text). We also developed ReID5o, a flexible framework that enables unified encoding and effective retrieval for any modality input. Our experiments demonstrate that leveraging multi-modal queries substantially improves retrieval performance compared to single-modality approaches, underscoring their practical value. We believe that the ORBench dataset and the ReID5o framework establish a solid foundation and a strong baseline for future research in this emerging field.

\section{Acknowledgments.}
This work was supported by the National Natural Science Foundation of China No.62176097, and the Hubei Provincial Natural Science Foundation of China No.2022CFA055. We would like to sincerely thank the HPC Platform of Huazhong University of Science and Technology for providing computational resources.
%%%%%%%%%%%%%%%%%%%%%%%%%%%%%%%%%%%%%%%%%%%%%%%%%%%%%%%%%%%%
\newpage
\bibliographystyle{plain}
\bibliography{main}

\newpage
\appendix
\section{Appendix}

\subsection{More Specific Samples}
We have shown and compared some specific typical samples of our proposed ORBench dataset and two existing tri-modal datasets in Fig~\ref{fig:supp}. As we can observe, due to the adoption of some rough data synthesis strategies to achieve the supplementation of additional modalities, the quality of existing tri-modal datasets is extremely low. For Tri-CUHK-PEDES~\cite{chen2023modalityagnostic}, since it is based on the existing CUHK-PEDES~\cite{cuhkpedes} and directly stylizes its RGB image data into a sketch style, we can notice that the sketch images in this dataset have a very low resolution and are significantly different in style from actual sketch drawings. For AIO~\cite{AIO}, it is based on the existing SYNTH-PEDES~\cite{PLIP} and directly uses some data augmentation methods to obtain the infrared and sketch modality. We can observe that the infrared and sketch images in this dataset are very rough and coarse-grained, failing to effectively cover the appearance characteristics of persons. 
However, for our proposed ORBench dataset, benefiting from meticulous manual annotation and repeated corrections, it possesses very high-quality five-modal data. The color pencil drawings and sketch drawings can maintain a high degree of identity consistency with the persons, and they have diverse painting angles and outstanding detail presentation capabilities. The text descriptions are extremely detailed and can be on a par with the existing finest text-based person ReID dataset UFine6926~\cite{zuo2024ufinebench}. 
Our ORBench, with its excellence in aspects such as the number of modalities, data diversity, and data quality, can make a very prominent contribution to the person ReID community and profoundly inspire subsequent researches.

\subsection{Public Evaluation Procedure}
In order to further evaluate the quality of our color pencil drawing data, we adopt the method of public evaluation to assess the identity consistency and perspective conformity. We randomly pick 2,000 samples from the color pencil drawings and split them into 20 groups. Then, 20 evaluators are asked to objectively score these samples on identity consistency with the original RGB images and conformity of drawing perspectives.

Regarding its identity consistency, the scores can be divided into five levels. The higher the score, the better the consistency. The meanings represented by each score are as follows:

\begin{figure}
    \centering
    \includegraphics[width=0.8\textwidth]{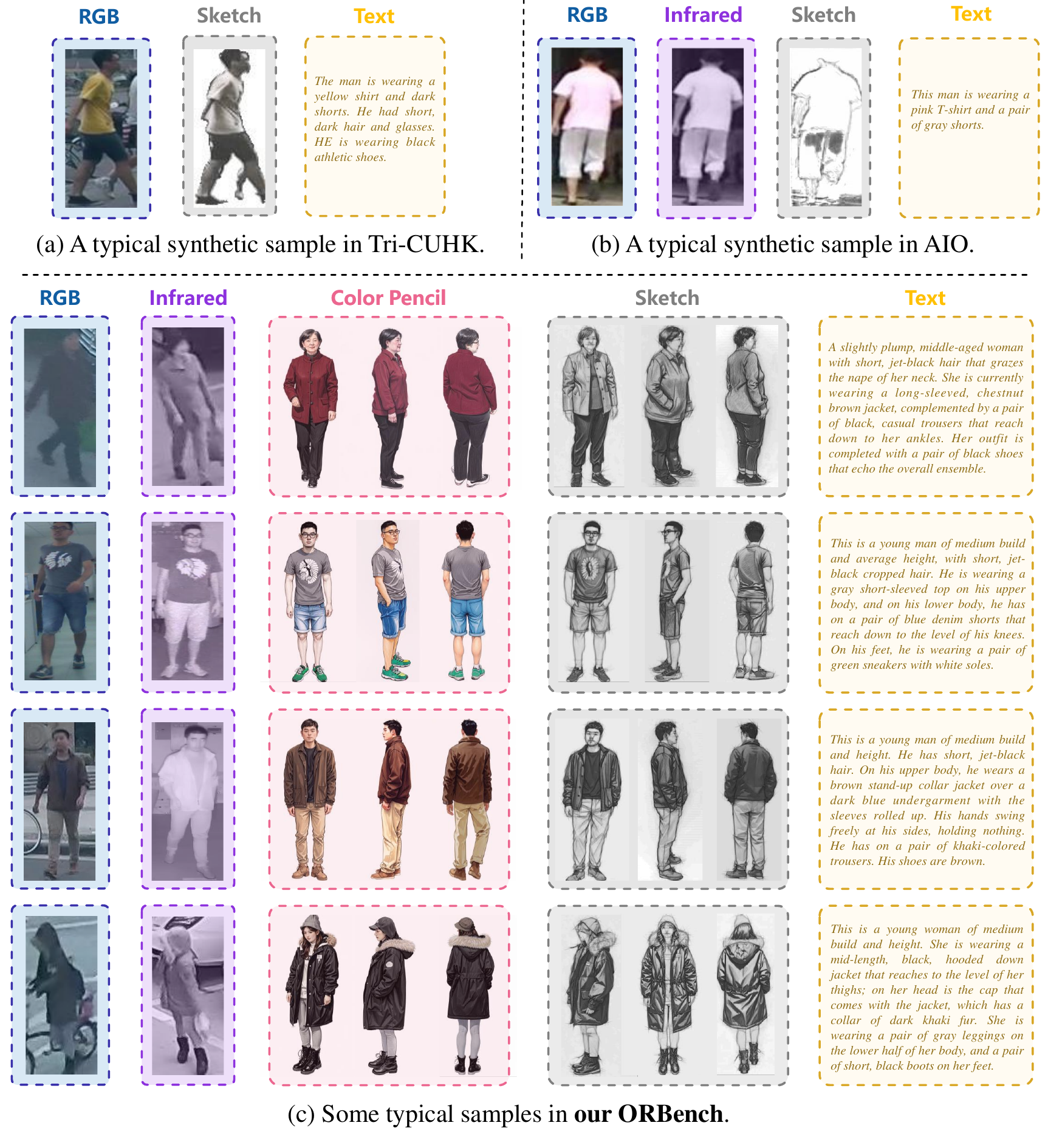}
    \vspace{-3mm}
    \caption{A visual overview of some typical samples in our proposed \textbf{ORBench} dataset and two existing multi-modal datasets~\cite{chen2023modalityagnostic,AIO}. As we can see, the quality of existing datasets is quite rough. 
    However, our ORBench not only encompasses a greater number of modalities but also demonstrates significant superiority in the data richness, quality and diversity.}
    \label{fig:supp}
    \vspace{-8mm}
\end{figure}

Score 1: There are two or more features in the color pencil drawing that are inconsistent with the original RGB image, such as the color of the shoes, whether wearing a hat or not, etc.

Score 2: There is one feature in the color pencil drawing that is inconsistent with the original RGB image.

Score 3: There are no features in the color pencil drawing that are inconsistent with the original RGB image, but there are relatively obvious painting errors, such as the orientation of the feet, etc.

Score 4: There are no features in the color pencil drawing that are inconsistent with the original RGB image, but there are few less obvious painting errors, such as the unnatural twisting of the hands, etc.

Score 5: There are no features in the color pencil drawing that are obviously inconsistent with the original RGB image, and there are no painting errors at all.

Regarding its perspective conformity, the scores can be divided into five levels. The higher the score, the better the conformity. The meanings represented by each score are as follows:

Score 1: The perspective does not match. The features of a certain perspective cannot be reflected in the color pencil drawing at all. For example, the label of the color pencil drawing is "back view", but the content of the drawing is actually a front view.

Score 2: The perspective partially does not match. The features of a certain perspective cannot be partially reflected in the color pencil drawing. For example, the label of the color pencil drawing is "front view", but the content of the drawing is a side view that shows only a small part of the front features.

Score 3: The perspective basically matches. The features of a certain perspective can be basically reflected in the color pencil drawing. For example, the label of the color pencil drawing is "front view", but the content of the drawing is a side view that basically shows the front features.

Score 4: The perspective mostly matches. The features of a certain perspective can be mostly reflected in the color pencil drawing. For example, the label of the color pencil drawing is "front view", but the content of the drawing is a side view that mostly shows the front features (such as turning slightly sideways about 10 degrees).

Score 5: The perspective completely matches. The features of a certain perspective can be completely reflected in the color pencil drawing. For example, if the label of the color pencil drawing is "front view", then the content of the drawing is a completely frontal image of the person.

\subsection{More Experimental Results}
\begin{table}[t]
\tiny
\centering
\caption{The performance of our ReID5o via the queries with different modality combinations. The fIst modality of each group is regarded as the primary modality. The best results are in bold.}
\resizebox{0.5\linewidth}{!}{
\begin{tabular}{l|ccccc}
\hline 
\multicolumn{1}{c|}{\multirow{2}{*}{Modality}} &\multicolumn{5}{c}{ORBench}\\
\cline{2-6}
 &Rank-1&Rank-5&Rank-10&mAP&mINP\\
\hline 
      I       & 52.288 & 69.170 & 75.674 & 43.644 & 16.694  \\
       C       & 87.222 & 95.139 & 96.903 & 75.093 & 37.599 \\
       S       & 69.833 & 85.389 & 89.944 & 58.719 & 23.935 \\
      T      & 63.147 & 77.939 & 82.920 & 54.884 & 19.053\\
     I+C     & 93.909 & 98.475 & 99.348 & 84.082 & 43.937 \\
     C+I     & 93.514 & 98.292 & 99.347 & 82.962 & 44.468 \\
     I+S     & 84.403 & 94.101 & 96.671 & 72.757 & 32.008 \\
     S+I     & 84.264 & 93.931 & 96.694 & 72.540 & 33.701 \\
    I+T    & 79.530 & 91.453 & 94.609 & 68.994 & 30.017 \\
    T+I    & 81.191 & 92.188 & 95.186 & 70.773 & 29.020 \\
     C+S      & 86.875 & 95.250 & 97.250 & 75.156 & 37.562 \\
     S+C      & 86.319 & 94.722 & 96.986 & 75.107 & 37.349 \\
    C+T     & 87.694 & 95.292 & 97.278 & 76.687 & 38.634 \\
    T+C     & 91.014 & 96.831 & 98.039 & 81.141 & 38.563 \\
    S+T     & 80.431 & 91.319 & 94.153 & 68.896 & 30.848\\
    T+S     & 84.709 & 93.391 & 95.546 & 74.005 & 30.868 \\
   I+C+S    & 95.127 & 99.070 & 99.693 & 85.028 & 44.344 \\
   C+I+S    & 94.778 & 98.875 & 99.597 & 83.941 & 45.096 \\
   S+I+C    & 95.153 & 98.778 & 99.583 & 83.983 & 45.127 \\
  I+C+T   & 95.031 & 98.839 & 99.597 & 85.655 & 45.728\\
  C+I+T   & 95.056 & 99.083 & 99.639 & 84.643 & 46.331 \\
  T+I+C   & 95.850 & 99.247 & 99.684 & 86.750 & 43.969 \\
  I+S+T   & 91.799 & 97.640 & 98.791 & 80.963 & 38.979 \\
  S+I+T   & 91.764 & 97.486 & 98.944 & 79.991 & 40.322 \\
  T+I+S   & 92.853 & 98.006 & 98.942 & 81.975 & 37.373 \\
   C+S+T   & 90.903 & 96.833 & 98.250 & 78.983 & 40.496 \\
   S+C+T   & 90.792 & 97.014 & 98.194 & 79.007 & 40.467 \\
   T+C+S   & 93.191 & 97.795 & 98.825 & 83.048 & 40.155 \\
 I+C+S+T & 95.751 & 99.127 & 99.645 & 86.501 & 46.628 \\
 C+I+S+T & 96.444 & 99.236 & 99.694 & 85.686 & 47.192 \\
 S+I+C+T & 96.000 & 99.083 & 99.694 & 85.742 & 47.156 \\
 T+I+C+S & 96.792 & 99.374 & 99.784 & 87.460 & 44.498 \\
    MM-1 Aver.    & 68.123 & 81.909 & 86.360 & 58.085 & 24.320 \\
    MM-2 Aver.    & 86.154 & 94.604 & 96.759 & 75.258 & 35.581 \\
   MM-3 Aver.   & 93.525 & 98.222 & 99.145 & 82.831 & 42.366 \\
   MM-4 Aver.    & 96.247 & 99.205 & 99.704 & 86.347 & 46.368 \\
\hline
\end{tabular}}
\label{supp_exp}
\end{table}

We have presented the retrieval results of our ReID5o in Tab~\ref{supp_exp}, for 32 different query sets on our ORBench dataset. It can be seen that the supplementation of each additional modality can effectively improve the retrieval performance. This inspires us that when actually deploying the ReID technology in real-world scenarios, we should focus on the development in the direction of multi-modal retrieval.

\subsection{Controlled Release}
To address the privacy concerns associated with the application of person ReID technology, we will implement a controlled release of our dataset and code, thereby preventing privacy violations and ensuring information security. We will require that users adhere to usage guidelines and restrictions to access our dataset and code. We have drafted the following regulations that must be adhered to, which will be refined and elaborated in subsequent releases: 

1. Privacy: All individuals using the ORBench dataset and ReID5o model should agree to protect the privacy of all the subjects in it. The users should bear all responsibilities and consequences for any loss caused by the misuse. 

2. Redistribution: The ORBench dataset and ReID5o model, either entirely or partly, should not be further distributed, published, copied, or disseminated in any way or form without a prior approval from the creators, no matter for profitable use or not.

3. Commercial Use: The ORBench dataset, ReID5o model, in full, part, or in derived formats, is not permitted for commercial use. Derivative works for commercial purposes are also prohibited. If a commercial entity wants to use them, a separate licensing agreement with the creators must be negotiated.

4. Modification: The ORBench dataset, in whole or part, cannot be modified. This includes changes to data elements, visual content, textual descriptions, and metadata, as well as structural changes. Modification is restricted to maintain dataset integrity for research. In rare cases where modification is needed, a request to the creators must be made, specifying the reasons and changes.

In parallel, we will require users to provide relevant information and will rigorously screen the submitted details to restrict access to our dataset and models by institutions or individuals with a history of privacy violations.

\subsection{Broad Impact Discussion}
This paper explores multi-modal person re-identification technology, which holds promise for applications in smart retail, intelligent transportation, and public safety systems by enabling cross-camera identification and tracking of individuals in urban environments.

However, the deployment of such technology must be approached with careful consideration of its ethical and societal implications. First, we explicitly acknowledge that the term “bias” in this context encompasses not only labeled sensitive attributes but also imbalances in data distribution. Our dataset, ORBench, is sourced primarily from university settings and does not reflect a globally uniform demographic distribution. Certain demographic groups—such as the elderly, individuals with visible disabilities, or specific racial groups—are likely underrepresented. As a result, model performance may vary across populations, and end-users should validate the fairness and generalization of the system in their specific application contexts prior to deployment.

Furthermore, while we adopted a structured annotation protocol to maximize objectivity and consistency—focusing on descriptive visual attributes such as clothing and accessories—we recognize that the design of such guidelines inherently introduces a form of methodological bias. By prescribing which features are salient, the annotation process shapes the model’s attention and may omit other potentially relevant characteristics.

The potential for misuse of this technology also warrants serious attention. Although our research is intended for academic and public safety purposes, the capability to retrieve individuals via textual descriptions could be repurposed to target people based on attributes associated with protected categories—such as disability, religious attire, or age. Like many dual-use technologies, person re-identification systems require strong governance, ethical oversight, and clear usage policies to prevent discriminatory or privacy-infringing applications. We oppose any such misuse.

In light of these considerations, we emphasize the importance of legal and regulatory frameworks to govern the application of ReID systems. Training often relies on surveillance data collected without explicit individual consent, raising privacy concerns. The research community should avoid using ethically problematic datasets and adopt responsible data practices. To mitigate potential harm, we will release ORBench under gated access, with usage restricted to non-commercial research purposes and subject to strict guidelines.

By addressing these issues transparently, we hope to encourage the responsible development and deployment of multi-modal person re-identification systems.

\end{document}